\documentclass{article}

\usepackage[final]{corl_2017}
\usepackage{graphicx}
\usepackage{floatrow}
\usepackage{subcaption}
\usepackage{caption}
\usepackage{subcaption}
\usepackage{array,booktabs}
\usepackage[final]{corl_2017}

\newfloatcommand{capbtabbox}{table}[][\FBwidth]

\title{Bodily aware soft robots: integration of proprioceptive and exteroceptive sensors}

\author{
  Gabor Soter\\
  Department of Engineering Mathematics\\
  University of Bristol,
  United Kingdom\\
  \texttt{Gabor.Soter@bristol.ac.uk} \\
  \And
  Andrew Conn\\
Department of Mechanical Engineering\\
University of Bristol,
United Kingdom\\
\texttt{A.Conn@bristol.ac.uk} \\
  \And
Helmut Hauser\\
Department of Engineering Mathematics\\
University of Bristol, 
United Kingdom\\
\texttt{Helmut.Hauser@bristol.ac.uk} \\
  \And
Jonathan Rossiter\\
Department of Engineering Mathematics\\
University of Bristol, 
United Kingdom\\
\texttt{Jonathan.Rossiter@bristol.ac.uk} \\
}
\usepackage{blindtext}
\usepackage{setspace}

\begin{document}

\maketitle

\begin{abstract}
Being aware of our body has great importance in our everyday life. This is the reason why we know how to move in a dark room or to grasp a complex object. These skills are important for robots as well, however, robotic bodily awareness is still an unsolved problem. In this paper we present a novel method to implement bodily awareness in soft robots by the integration of exteroceptive and proprioceptive sensors. We use a combination of a stacked convolutional autoencoder and a recurrent neural network to map internal sensory signals to visual information. As a result, the simulated soft robot can learn to \textit{imagine} its motion even when its visual sensor is not available.
\end{abstract}

\keywords{soft robots, bodily awareness, proprioception} 


\section{Introduction}
Bodily awareness is essential for human beings. We use it to understand the state of our body, to be aware of its position, movement, and to process sensations, such as pain or temperature. The sense of body-ownership is produced by the integration of a wide range of sensors \cite{intro}. These provide information either on the outside world (exteroceptive signals), such as our vision and touch sensors, or on the state of the inner body. The latter can be further divided into two groups: proprioceptive sensors, e.g. muscle spindles, that we use to sense our body position and movement, and interoceptive sensors, that allows us to monitor our internal body, e.g. blood pressure or hunger. Recent studies have shown that the brain integrates these signals and interoceptive or proprioceptive awareness can modulate external representations \cite{Tsakiris2011JustAH}, \cite{proprio}.

The first of these representations are developed in the fetal stage \cite{fetal}. Over the early course of our lives bodily awareness evolves as our brain learns to map the proprioceptive and exteroceptive signals, and this mapping plays a crucial role later in accomplishing any task in which the body is involved. Given the importance of this integration and its obvious usefulness in robotics, several groups addressed the problem of robotic bodily awareness, since its underlying mechanisms are co-responsible for controlling complex bodies, adapting to growth, and using tools \cite{robot1}, \cite{hosoda}. Most of these implementations used conventional robots, that have hard components, relatively simple and constrained behaviour and can be modelled with few degrees of freedom \cite{review}. However, the complex bodies of a new class of robots known as soft robots introduce many challenges and it remains an open question how soft robots can develop bodily awareness. In this paper we try to address this, more specifically, we propose an approach that integrates exteroceptive and proprioceptive signals for these new types of robot.

Soft robots are flexible, compliant and predominantly made of elastic materials. They have very high number of degrees of freedom and their movement is complex. They are often underactuated \cite{naka}, \cite{uact}, which means that their control is not trivial. In the past few years data driven, model-free control methods have gained great interest due to their ability to learn from data and to approximate nonlinear, hyperelastic behaviour \cite{barry}, \cite{romi}. This usually exploits artificial neural networks to create generalised models. In this work our goal is to use model-free techniques to describe the body and the motion of a soft octopus arm \cite{octopus} and to implement bodily awareness by the integration of proprioceptive and exteroceptive information.

The proprioceptive representation in our setup is estimated by stretch sensors located inside of the soft body. This is a biologically inspired approach: a wide variety of species including the octopus have stretch sensors around their muscles \cite{animaals} that are used for self sensing.
The exteroceptive representation is based on visual data, whose dimensionality can be reduced by encoding techniques. The advantages of this low-dimensional representation is that it can be mapped to temporal signals of other sensors, allowing the robot, for example, to \textit{imagine} its body even if the visual sensor is not available.

\section{Methodology}
\label{sec:citations}
\begin{figure}[H]
		\begin{subfigure}{.5\textwidth}
			\centering
			\hskip-0.5cm
			\includegraphics[width=.205\linewidth]{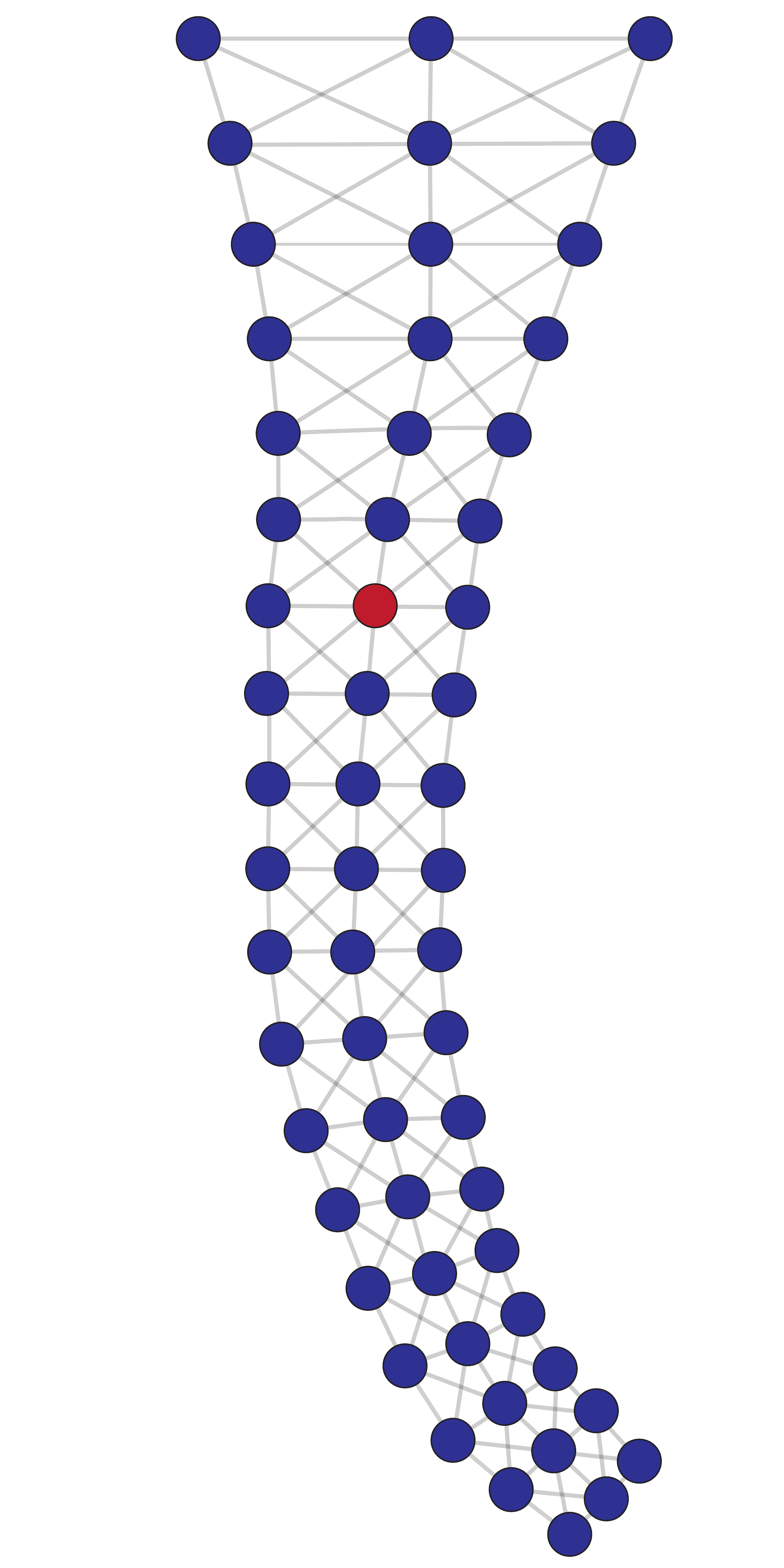}
		  	\caption{}
			\label{fig:sfig1}
		\end{subfigure}%
		\begin{subfigure}{.5\textwidth}
			\centering
			\includegraphics[width=.65\linewidth]{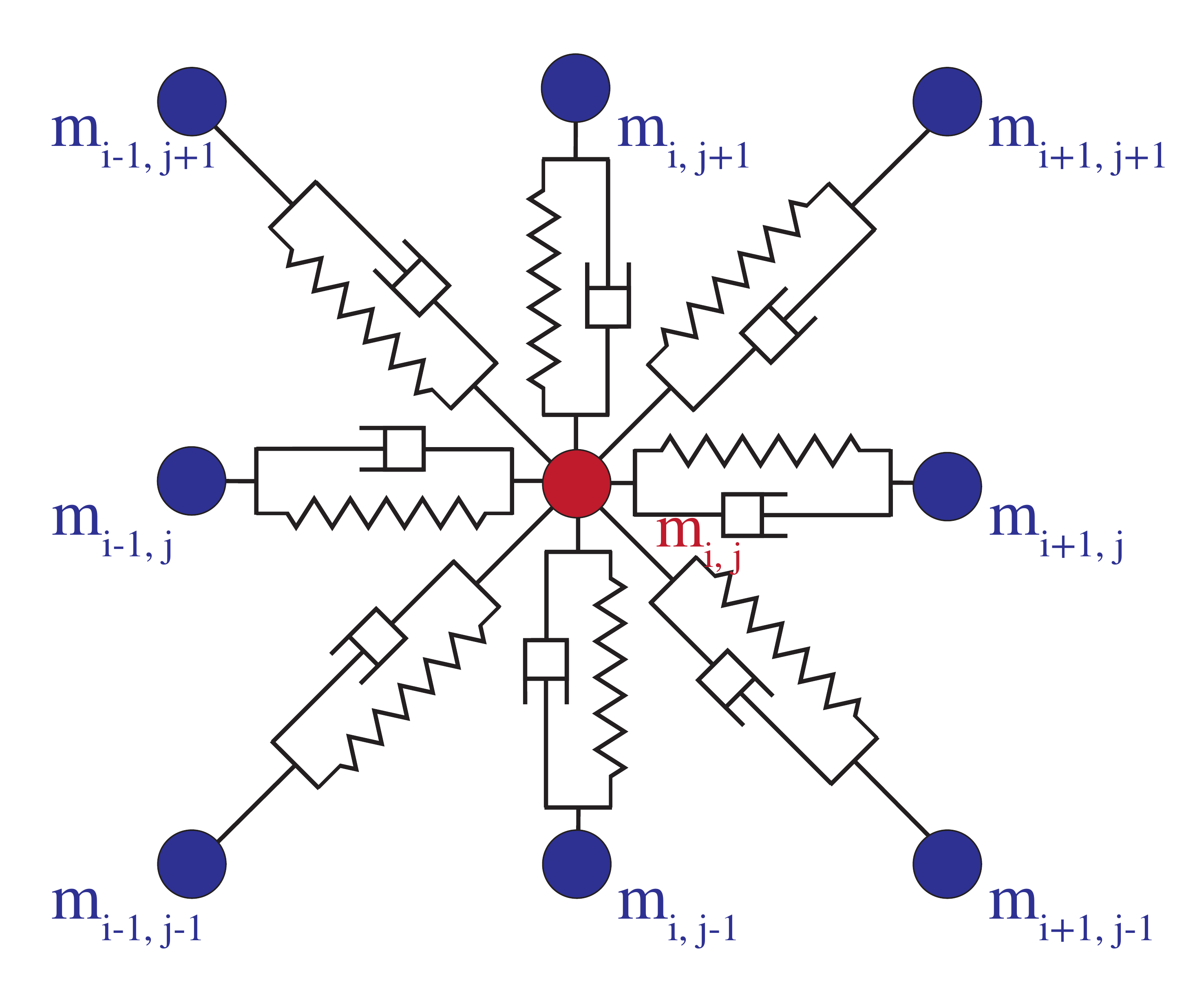}
			\caption{}
			\label{fig:sfig2}
		\end{subfigure}
		\caption{(a) The octopus arm was modelled by a two-dimensional mass-spring damper system \cite{naka}, where the neighbouring masses were connected by viscous-elastic elements. A typical node is shown with its connections in (b). The model included 57 nodes with the mass $m = 0.01$ kg, all springs had linear characteristics with the stiffness $k = 1$  kN/m and all viscous dampers had the same damping coefficients $b = 0.9$ Ns/m. Each sensor was connected to two nodes and the stretch was defined as the change of the Eucledian distance between the two nodes.}
		\label{msd}
	\end{figure}

	We developed a soft robot simulation framework that is able to model a generic two-dimensional continuum body. The simulation is based on a two dimensional mass-spring-damper system, where the neighbouring masses are connected by viscous-elastic elements (Figure \ref{msd}). The physical model is described by a system of ordinary differential equations, which can be solved numerically. In order to produce complex, nonlinear and non-periodic behaviour, the top of the soft body was excited (horizontally moved) by a product of sinusoidal functions
	\begin{equation}
	f(t) =  A\cdot sin ( 2 \pi f_{1} t) \cdot sin(2 \pi f_{2} t) \cdot sin(2 \pi f_{3} t) ,
	\end{equation}
	where $A$ is the amplitude, $f_{1} =  2.11$ Hz, $f_{2} = 3.73$ Hz and $f_{3} = 4.33$ Hz are the frequencies of excitation. Choosing these particular frequency values ensures a long enough time period for $f(t)$ ($T = 100$ s) to generate sufficient data points for learning and evaluation without repeating the same signal \cite{Hauser2011}.
	
\begin{figure}[H]
	\begin{subfigure}{.5\textwidth}
		\centering
		\includegraphics[width=1\linewidth]{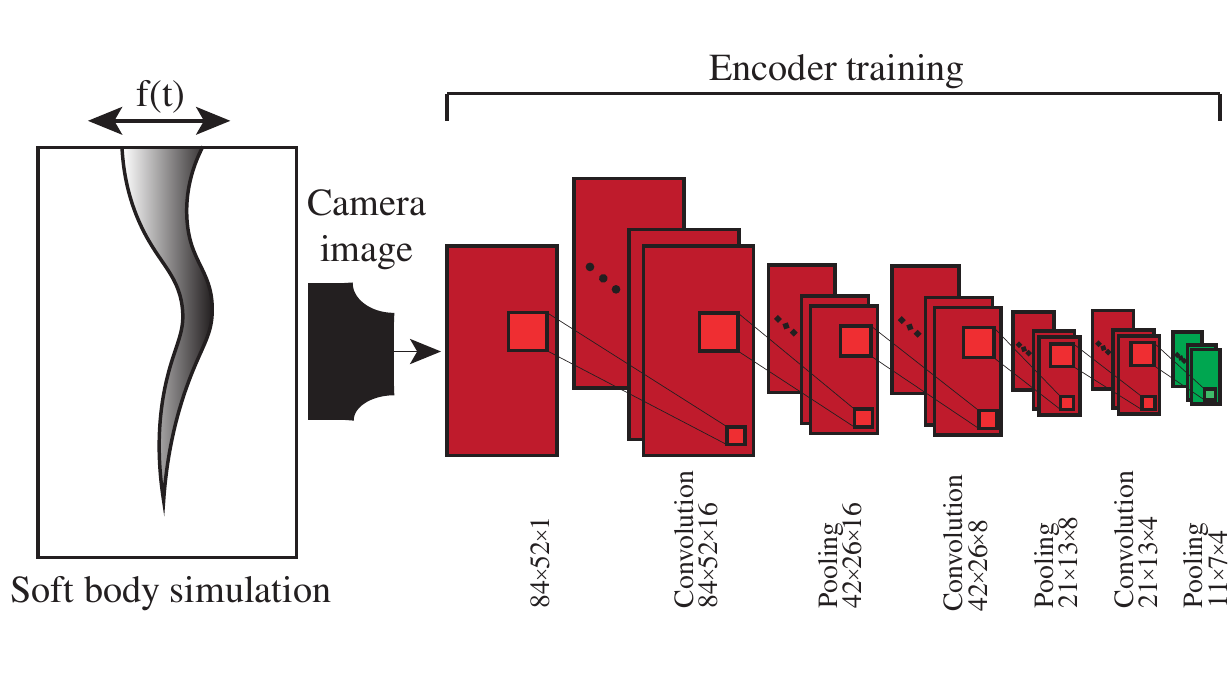}
		\caption{}
		\label{}
	\end{subfigure}%
	\begin{subfigure}{.5\textwidth}
		\centering
		\includegraphics[width=1\linewidth]{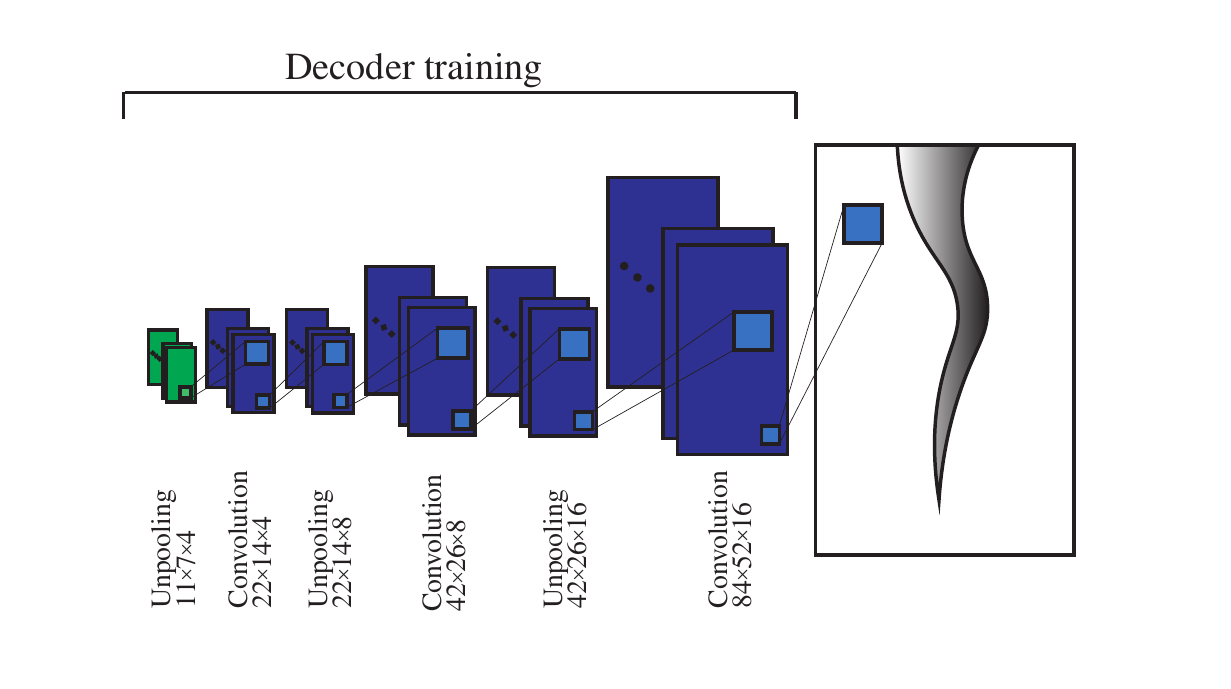}
		\caption{}
		\label{}
	\end{subfigure}
\vskip0.8cm
\begin{subfigure}{1\textwidth}
	\centering
	\includegraphics[width=0.9\linewidth]{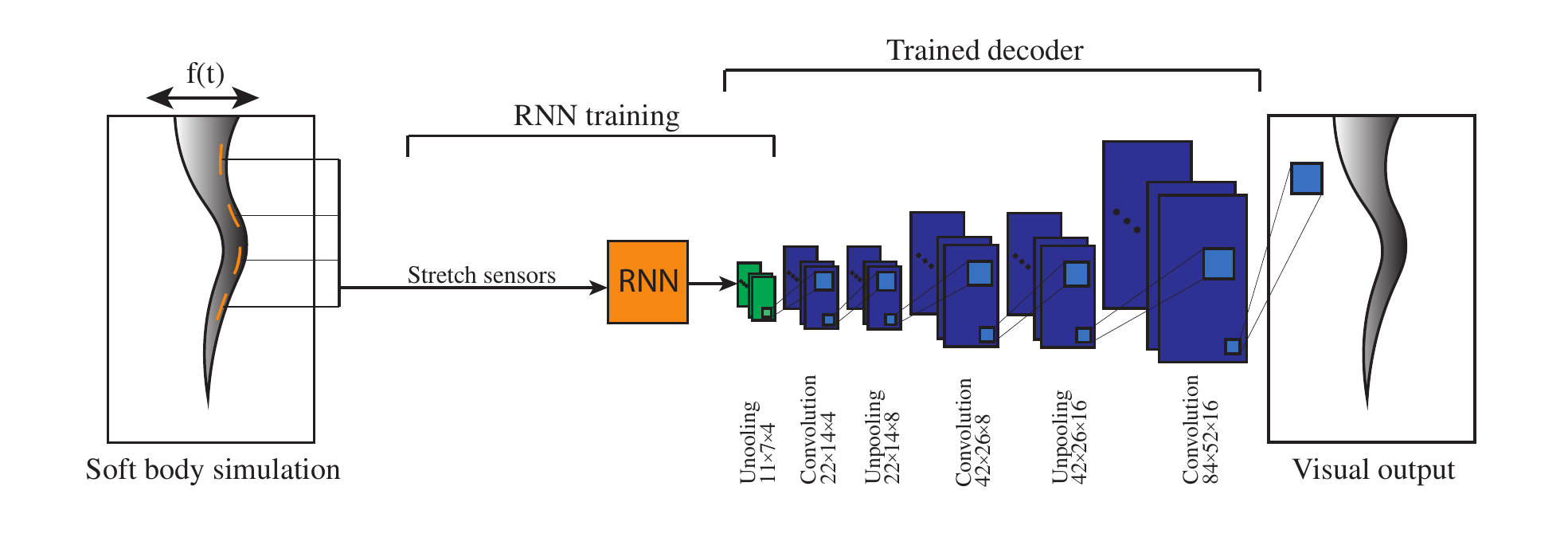}
	\caption{}
	\label{}
\end{subfigure}
	\caption{The system architecture and the learning process. First, we used a stacked convolutional model to reduce the dimensionality of the exceroceptive (visual) representation of the body. Both the encoder (a) and decoder (b) components had five hidden layers.  (c) After autoencoder training, we mapped the temporal data received from stretch sensors to this low-dimensional encoded representation (shown in green). Finally, by attaching the previously trained decoder component the robot could \textit{imagine} its body even when the visual sensor was not available.}
	\label{arch}
\end{figure}

In order to make the robot learn to \textit{imagine} its body using proprioceptive stretch sensors, the temporal signals of the stretch sensors were mapped to the visual model of the simulation. Our goal was to train a single layer, long short-term memory (LSTM) type of recurrent neural network (RNN) using data from four stretch sensors as inputs and the complete video frames of the simulation as outputs. However, in this case the number of neurons on the output layer (the number of pixels in each image, 4368 in this study) was three orders of magnitude higher than on the input layer and the trained recurrent neural network was underfit. Therefore, we used a stacked convolutional autoencoder \cite{Masci2011} in order to decrease the dimensionality of the visual images and to be able to train the RNN with reasonable accuracy. The architecture of the proposed system and the learning process is shown in Figure \ref{arch}. First, the stacked convolutional encoder and decoder pair was trained on the visual images in order to find a low-dimensional representation of the soft body's movement. Next, these features and the stretch sensors' data were used to train the recurrent neural network. Finally, the trained decoder component was connected to the RNN, creating a path for the information flowing from the stretch sensors to the visual output.

\section{Experimental Results}
\label{sec:result}
We trained and tested the system using the data produced by a 100 s long simulation with the time step t = 0.01 s. We used the first 85 s (8500 images) to train and the next 10 s (1000 images) to test the stacked convolutional autoencoder. All the images were binary black and white and contained 84$\times$52 pixels that were normalized before the training. In order to measure the error of the trained model, we calculated the binary cross-entropy testing loss between the original and reconstructed images, which converged to 0.119. 

After this, we trained the recurrent neural network with the same dataset to map the temporal data received from the stretch sensors to the previously encoded representation. A high pass filter was used to remove the small number of encoded features that did not change while the body was moving. Both the encoded features and the stretch sensors' data were normalized prior to the training, and we used the previous six data points to predict the next value of the signals.

\begin{figure}[H]
	\begin{subfigure}{0.975\textwidth}
		\centering
		\hspace*{-0.35cm}
		\includegraphics[width=0.86\linewidth]{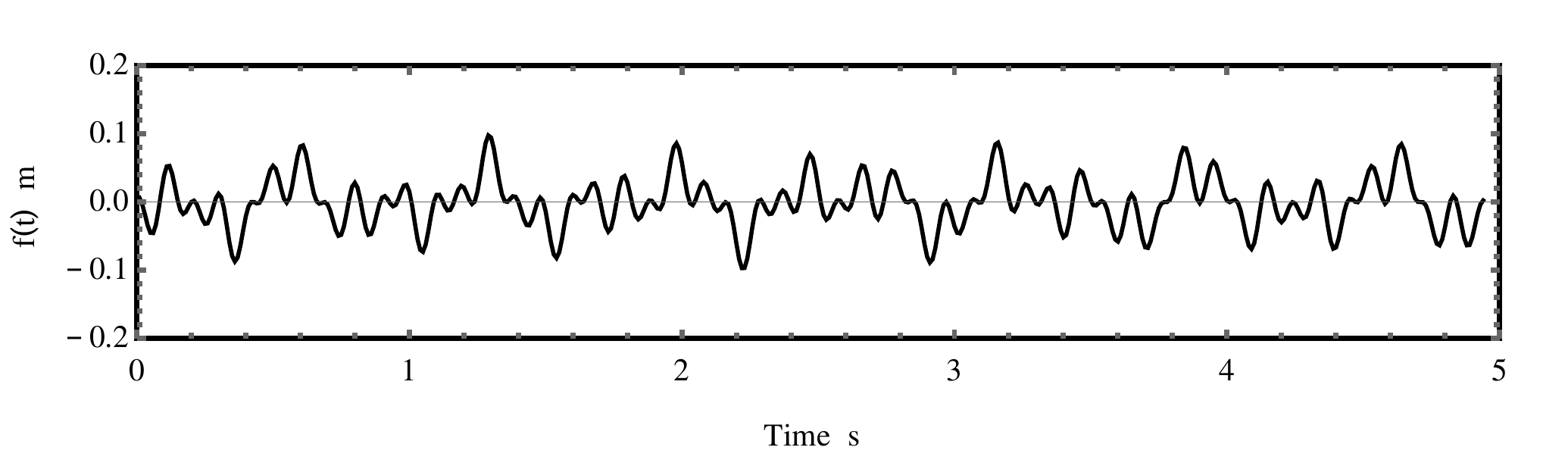}
		\caption{}

	\end{subfigure}
\begin{subfigure}{1\textwidth}
	\centering
	\includegraphics[width=0.8\linewidth]{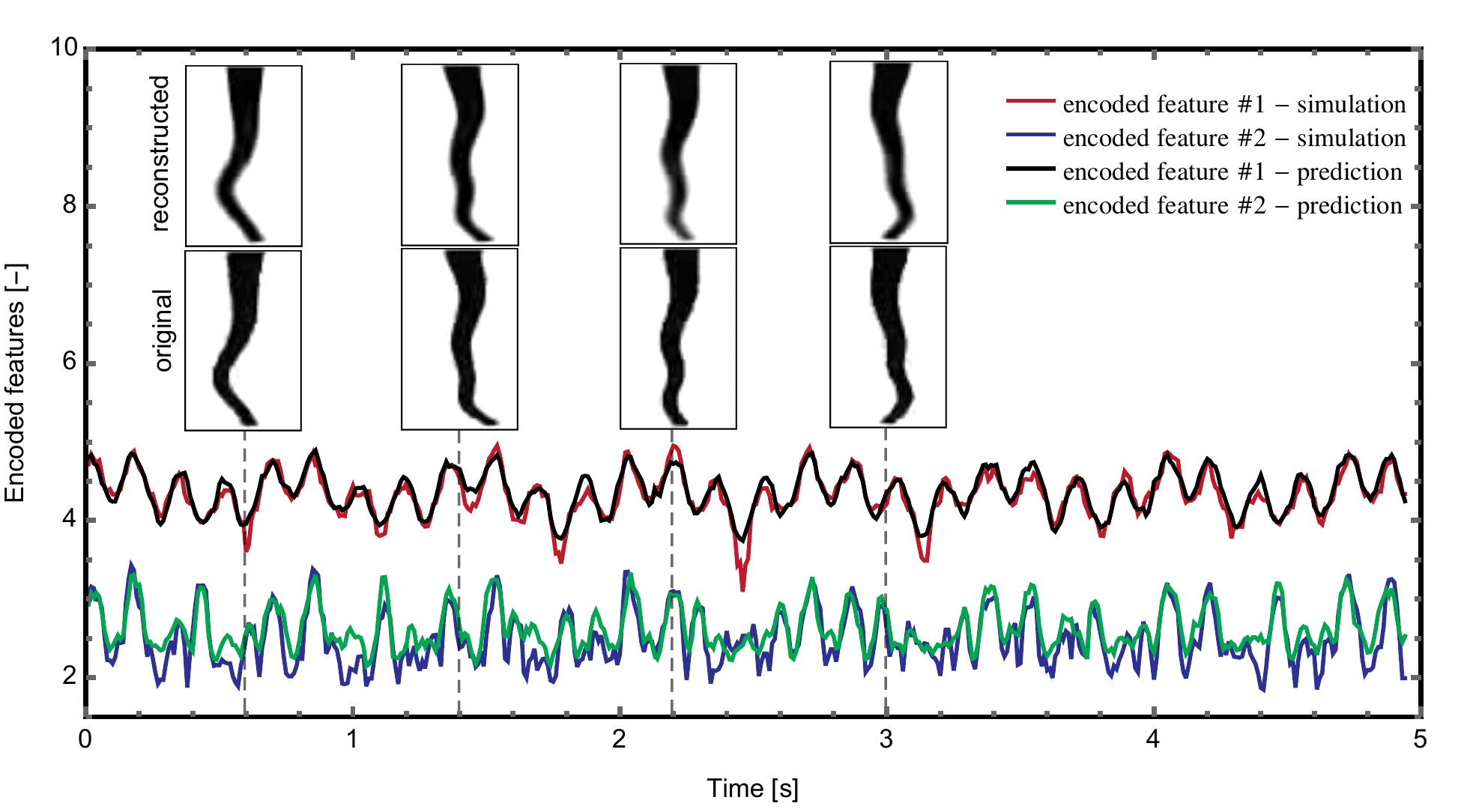}
	\caption{}

\end{subfigure}

\caption{(a) Excitation function over time. The plot shows how the excitation changes over 494 data points. (b) Comparison between the simulated and predicted signals of two encoded features. The figure also shows the original and reconstructed images. We note that these data points have not been used before for training.}
\label{ha}
\end{figure}

The encoded features predicted by the recurrent neural network are shown in Figure \ref{ha}. We reached 0.051 cross-entropy testing loss. As is shown in Figure \ref{arch}, these were fed into the previously trained decoder and the images representing the soft body could be reconstructed. We note that this was achieved using the data of the last 5 s of the simulation, that have not been used before for training. Since the recurrent neural network predicts the next value of the signal using the previous six data points, we used the remaining 494 to generate the plot.

\section{Conclusion}
\label{sec:conclusion}
In this paper we implemented bodily awareness in a soft robot by the integration of exteroceptive and proprioceptive sensors. We developed a two dimensional mathematical model that is able to simulate the nonlinear motion of an octopus-inspired arm. The proprioceptive data were collected from stretch sensors located inside the body, while for exteroceptive representation we used visual data, whose dimensionality was reduced by a stacked convolutional autoencoder. We connected and trained a recurrent neural network to the previously trained decoder component and mapped the temporal data received from the proprioceptive sensors of the robot to the encoded features. Using this system the robot could \textit{imagine} its body moving. The advantage of the proposed architecture is that it can be used for complex, non periodic and nonlinear movements produced by soft robotic bodies. In the future we expect our method to be used to implement bodily awareness into real robots and to improve their efficiency in all kinds of tasks in which the body is involved, e.g. locomotion in uncertain environments and manipulation of complex objects.

\section{Acknowledgements}
This research was partially funded by EPSRC grants EP/M026388/1 and EP/M020460/1 and Leverhulme Trust research project RPG-2016-345.





\bibliography{example}  

\end{document}